\documentclass{article}

\usepackage[preprint,nonatbib]{confstyle}

\usepackage[utf8]{inputenc} 
\usepackage[T1]{fontenc}    
\usepackage{hyperref}       
\usepackage{url}            
\usepackage{booktabs}       
\usepackage{amsfonts}       
\usepackage{nicefrac}       
\usepackage{microtype}      
\usepackage{xcolor}         
\usepackage{graphicx}  
\usepackage{setspace}
\usepackage{multirow}

\usepackage{mathptmx} 
\usepackage{times} 
\usepackage{amssymb}  

\usepackage{listings} 

\usepackage{algorithm}  
\usepackage{algorithmic}
\usepackage{amsmath}

\usepackage{balance}
\usepackage{adjustbox}
\usepackage{tabularx}

\title{\Large TeachAnything: A Multimodal Crowdsourcing Platform for Training Embodied AI Agents in Symmetrical Reality }

\author{
\textbf{Zidong Liu} \quad
\textbf{Rongkai Liu} \quad
\textbf{Yue Li} \quad
\textbf{Zhenliang Zhang}\thanks{E-mail: zlzhang@bigai.ai}\\ 
State Key Laboratory of General Artificial Intelligence, BIGAI
}

\makeatletter
\def\thanks#1{\protected@xdef\@thanks{\@thanks
        \protect\footnotetext{#1}}}
\makeatother

\begin{document}

\maketitle

\begin{abstract}
   Symmetrical Reality (SR) is emerging as a future trend for human–agent coexistence, placing higher demands on agents to acquire human-like intelligence. It calls for richer and more diverse human guidance. We introduce a three-stage demonstration paradigm integrating multimodal demonstration signals. Building on this paradigm, we developed \textbf{TeachAnything}, a cloud-based, crowdsourcing-oriented demonstration platform with physics simulation capable of collecting diverse demonstration data across varied scenes, tasks, and embodiments. By unifying virtual and physical interactions through both methodological design and physics simulation, the system serves as a practical foundation for developing embodied agents aligned with Symmetrical Reality.
\end{abstract}

\section{Introduction}

Symmetrical Reality (SR) is increasingly regarded as an inevitable developmental trend in embodied AI, envisioning a future where physical and virtual worlds integrate seamlessly, and intelligent agents interact coherently across both domains.~\cite{zhang2024emergence} Achieving such embodied intelligence requires agents to develop unified perception and action capabilities across realities, which fundamentally depends on large-scale, diverse, and semantically aligned demonstrations.~\cite{yifan2025embodied} Moreover, modern human-in-the-loop learning frameworks, including VLA models and policy learning systems, also depend on large-scale multimodal data to bridge high-level intent, perceptual grounding, and low-level control.~\cite{zitkovich2023rt}
However, existing demonstration pipelines are misaligned with the needs of this SR-driven future. They are often restricted to fixed scenarios, predefined tasks, single embodiments, or single-modality inputs, limiting the richness and variability of supervision they can provide.~\cite{mandlekar2018roboturk} Complex real-world tasks commonly require multimodal teaching signals, span a wide variety of environments and goals, and often involve the generation of complex demonstration data, such as manipulation tasks that produce continuous action trajectories, which pose higher technical challenges for existing demonstration methods. As a result, a substantial gap emerges between the data required to train SR-capable agents and what existing demonstration-collection systems can provide.

To address these limitations, we introduce a three-stage demonstration paradigm that explicitly targets the key challenges in SR-oriented data collection. By decomposing human teaching into semantic, perceptual, and embodied channels, the paradigm enables multimodal supervision for complex tasks, supports open-ended demonstrations across diverse scenes and tasks, and provides dedicated mechanisms for generating fine-grained continuous action data. Specifically, language demonstrations capture high-level intent and task structure beyond fixed templates. Video demonstrations ground task execution by providing rich perceptual evidence from diverse scenes and embodiments, and teleoperation-based demonstrations produce precise continuous control trajectories for manipulation-intensive skills. These stages form a coherent and scalable framework that bridges the gap between the rich demonstration data required by SR-capable agents and the limitations of existing demonstration pipelines.

Building on this paradigm, we develop \textbf{TeachAnything}, a cloud-based demonstration platform that unifies all three stages within a single environment. The system supports configurable scenes, multiple interaction channels, and diverse robot embodiments. An Isaac Sim backend powered by PhysX provides high-fidelity physical interactions for embodiments such as the Franka arm and Unitree G1, while WebSocket streaming synchronizes scenes and commands, and Flask microservices enable camera input and HaMeR-based~\cite{pavlakos2024reconstructing} gesture control. All language, video, and teleoperation data are consolidated into a unified structured format for embodied-agent training and virtual–physical integration in Symmetrical Reality.

\begin{figure}[t]
  \centering
  \includegraphics[width=\linewidth]{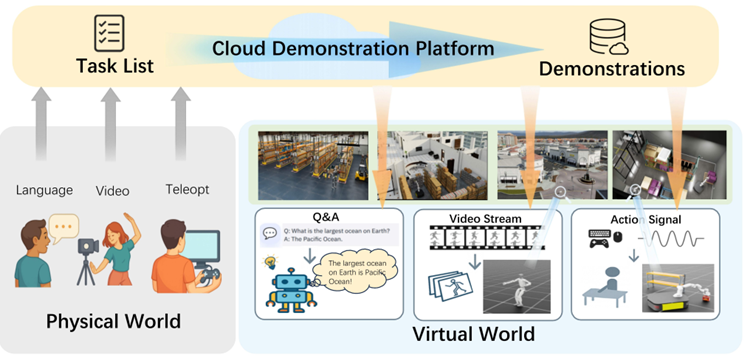}
  \vspace{5pt}
  \caption{\textbf{Overview of TeachAnything.} A cloud-based crowdsourcing demonstration platform that enables users to teach anytime and anywhere through multimodal demonstrations. The platform supports both predefined and user-defined tasks within rich virtual scenes, and converts heterogeneous inputs into structured data for training embodied agents.}
  \label{fig:teaser}
\end{figure}

\section{Methods}
Our methodology centers on a three-stage paradigm that organizes human guidance by modality and interaction channel. Each stage serves as an information-carrying pathway for human demonstration, independent of specific sensors, embodiments, or task settings. By decoupling the demonstration interface from the underlying agent or environment, this structure unifies heterogeneous demonstration sources and enables consistent interpretation across tasks, robot embodiments, and virtual–physical domains.

\begin{itemize}
    \item \textbf{Language-based demonstration}: Text or speech inputs that describe task execution, articulate high-level goals, or provide semantic annotations of demonstration content. This modality not only conveys procedural intent but also supplies contextual cues, such as object relations, constraints, or the rationale behind certain actions that may not be visually observable.

    \item \textbf{Video-based demonstration}: Uploaded or recorded videos of complete task executions or annotated visual traces, covering human demonstrations, robot executions, and simulation renderings across diverse scenes and embodiments. Videos provide temporally dense supervision of motions and interactions, supporting the learning of spatial reasoning, action dynamics, and visual affordances from observation.

    \item \textbf{Teleoperation-based demonstration}: Real-time control of an embodied agent in simulation, producing continuous action trajectories via various interfaces such as keyboard-mouse input and vision-based gesture control. This modality provides fine-grained motor supervision aligned with the control loop, enabling the collection of detailed manipulation strategies and corrective refinements for training low-level policies.
\end{itemize}

\section{System Demostration}

\begin{figure}[t]
  \centering
  \includegraphics[width=\linewidth]{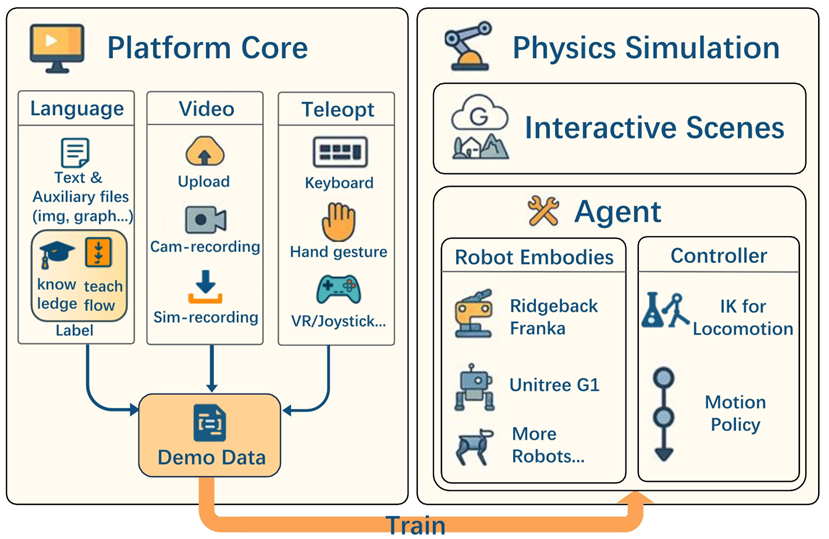}
  \vspace{-10pt}
  \caption{\textbf{System pipeline of TeachAnything.} The system integrates language, video and teleoperation input channels with a physics-based simulator and a WebSocket streaming layer, producing temporally aligned training data for SR-aligned embodied agents.}
  \label{fig:system}
\end{figure}

To realize the proposed paradigm, we implement a cloud-based crowdsourcing platform that enables users to initiate demonstrations anytime and anywhere for both predefined and user-defined tasks, shown in Fig~\ref{fig:system}. The platform provides a unified workflow to collect language, video, and teleoperation demonstrations within a single simulation environment, with all data logged in a synchronized format.

\begin{figure}[t]
  \centering

  \begin{minipage}{\linewidth}
    \centering
\includegraphics[width=\linewidth,height=0.35\textheight,keepaspectratio]{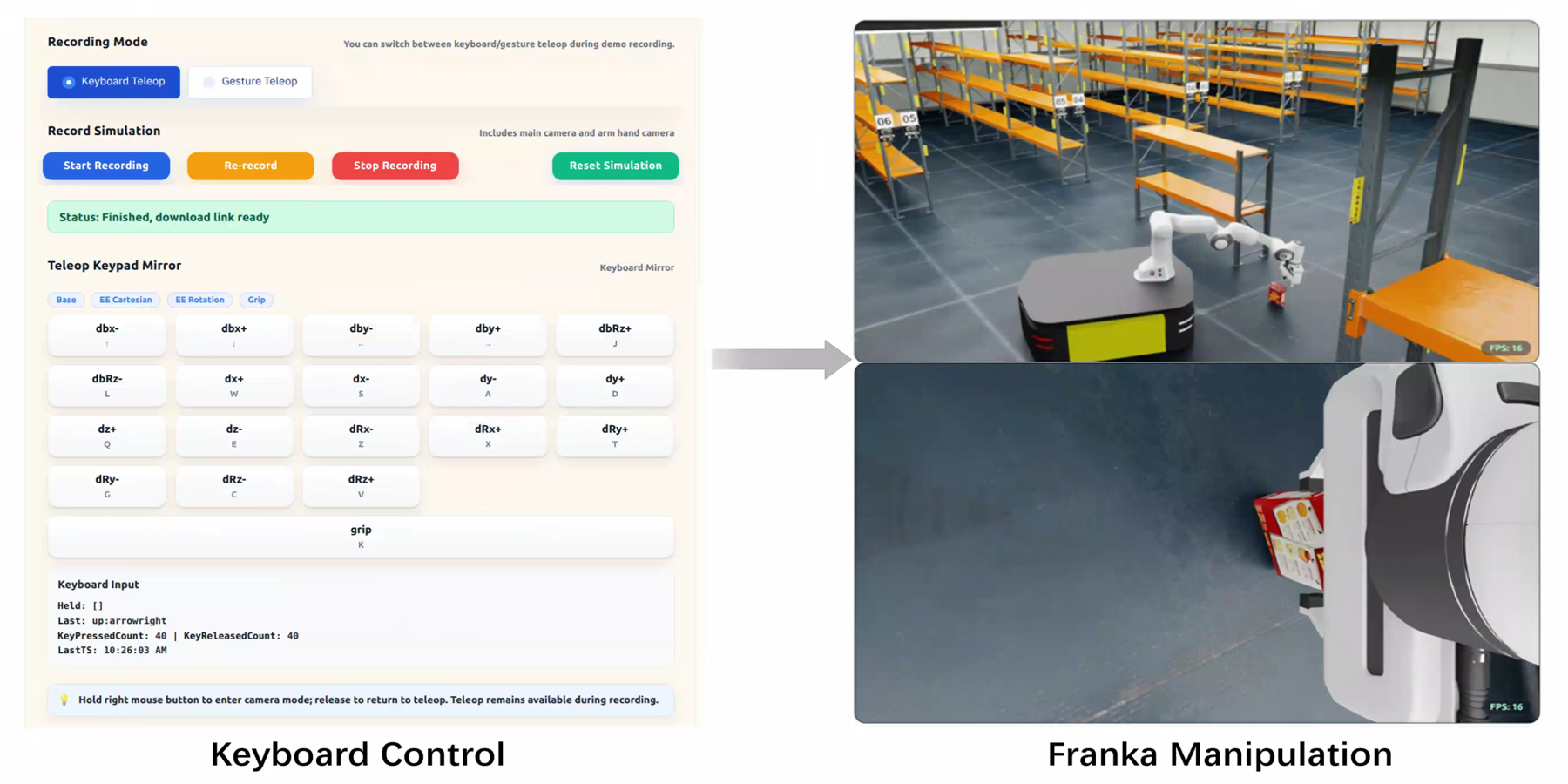}
  \end{minipage}

  \begin{minipage}{\linewidth}
    \centering
    \includegraphics[width=\linewidth,height=0.35\textheight,keepaspectratio]{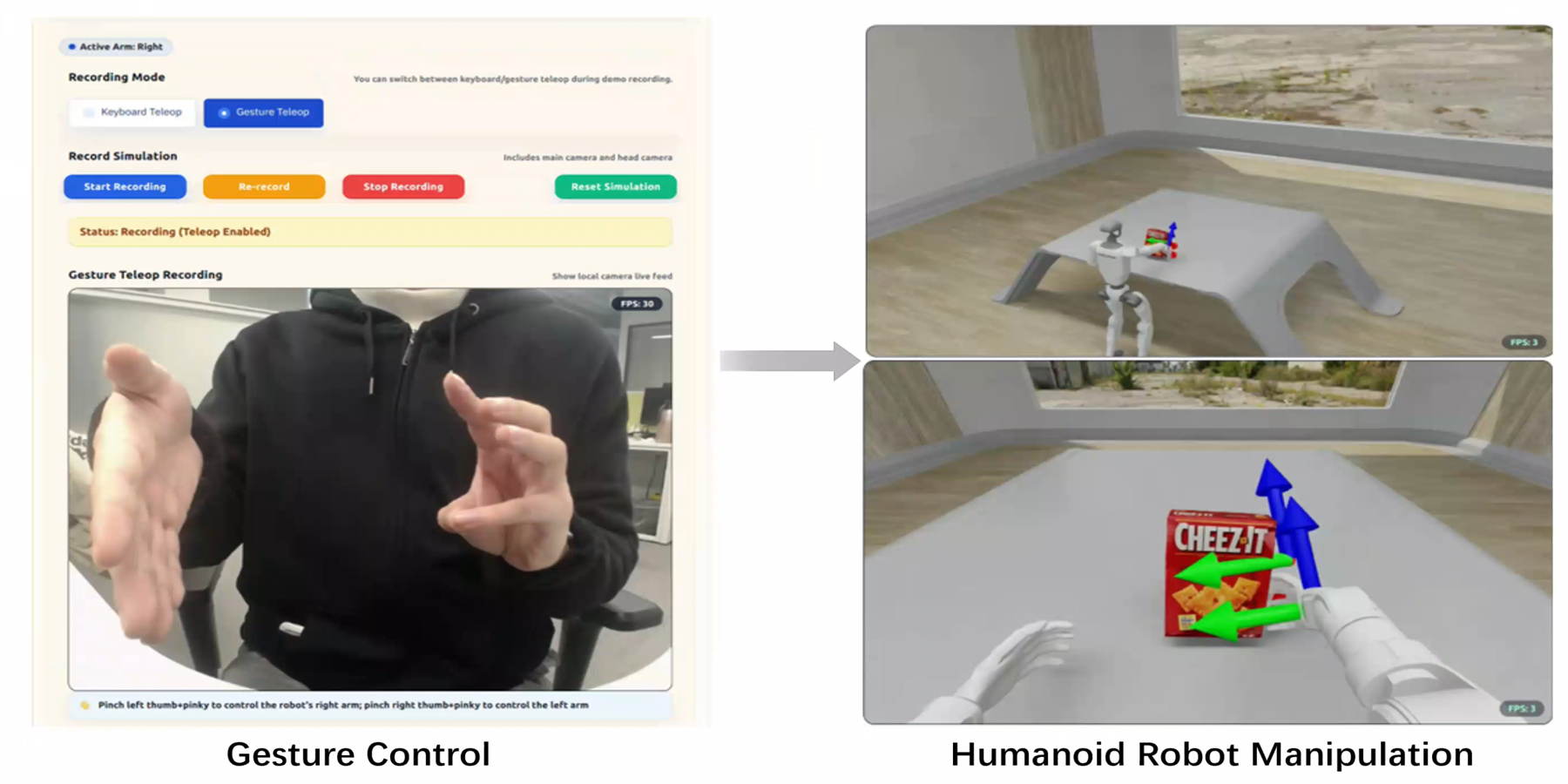}
  \end{minipage}
  \vspace{5pt}
  \caption{Example teleoperation demonstrations collected on the platform: keyboard-mouse control of a Franka arm and vision-based hand-gesture control of a Unitree G1.}
  \label{fig:platform}
\end{figure}

The platform core orchestrates multimodal teaching from distributed users. Language demonstrations allow users to describe tasks, procedures, and constraints in free-form text, which are automatically organized with semantic metadata. Video demonstrations support uploads, local camera recording, and simulation capture, allowing reset and re-recording, and each video is stored with structured metadata to facilitate cross-modal alignment. For embodied supervision, the platform supports teleoperation via keyboard–mouse input and vision-based hand-gesture control, through which continuous action commands and synchronized simulation states are streamed to enable low-latency interaction (Fig.~\ref{fig:platform}).

All demonstrations run on a physics-based simulation backend with diverse robot embodiments and interactive scenes. Robots (e.g., Franka and Unitree G1) are controlled by dedicated stacks that combine inverse kinematics and learned motion policies to generate physically consistent demonstrations.

\section{Conclusion And Future Work}

We present a three-stage demonstration paradigm and a cloud-based crowdsourcing platform that support scalable, multimodal supervision for training embodied agents in Symmetrical Reality. By unifying language, video, and teleoperation demonstrations within a physics-based simulation environment, the platform enables flexible and open-ended teaching across tasks, scenes, and embodiments. Although the platform is still evolving, we plan to integrate VR-based teleoperation, demonstrate SR/VR interaction with embodied agents, and complete an end-to-end data-to-training pipeline, along with pilot user studies to assess usability and data quality across different interaction channels, thereby further strengthening scalable supervision for SR-aligned embodied learning.



\end{document}